\documentclass[sigconf]{acmart}
\usepackage{balance} %
\AtBeginDocument{%
  \providecommand\BibTeX{{%
    \normalfont B\kern-0.5em{\scshape i\kern-0.25em b}\kern-0.8em\TeX}}}

\DeclareUnicodeCharacter{202F}{FIX ME!!!!}

\usepackage{amsmath} 
\usepackage{pgfplots} 
\pgfplotsset{compat=1.8} 
\usepackage{graphicx} 
\usepackage{xcolor} 
\usepackage{tikz}
\usepackage{caption}
\usepackage{subcaption}
\usepackage{pgf-pie}  
\usepackage{geometry}
\begin{document}
\title{Leveraging Large Language Models in Human-Robot Interaction: A Critical Analysis of Potential and Pitfalls}
\author{Jesse Atuhurra}
\orcid{0000-0002-3717-8924}
\affiliation{%
  \institution{Nara Institute of Science and Technology}
  \city{Ikoma}
  \state{Nara}
  \country{Japan}
  \postcode{630-0101}
}
\email{atuhurra.jesse.ag2@.naist.ac.jp}

\begin{abstract}
The emergence of large language models (LLM) and, consequently, vision language models (VLM) has ignited new imaginations among robotics researchers. At this point, the range of applications to which LLM and VLM can be applied in human-robot interaction (HRI), particularly socially assistive robots (SARs), is unchartered territory. However, LLM and VLM present unprecedented opportunities and challenges for SAR integration.
We aim to illuminate the opportunities and challenges when roboticists deploy LLM and VLM in SARs. 
First, we conducted a meta-study of more than 250 papers exploring 1) major robots in HRI research and 2) significant applications of SARs, emphasizing education, healthcare, and entertainment while addressing 3) societal norms and issues like trust, bias, and ethics that the robot developers must address. 
Then, we identified 4) critical components of a robot that LLM or VLM can replace while addressing the 5) benefits of integrating LLM into robot designs and the 6) risks involved.  
Finally, we outline a pathway for the responsible and effective adoption of LLM or VLM into SARs, and we close our discussion by offering caution regarding this deployment. 
\end{abstract}


\maketitle

\section{Introduction}
\label{sec:Introduction}
\begin{figure}[!t]
  \centering
  \includegraphics[width=0.95\linewidth]{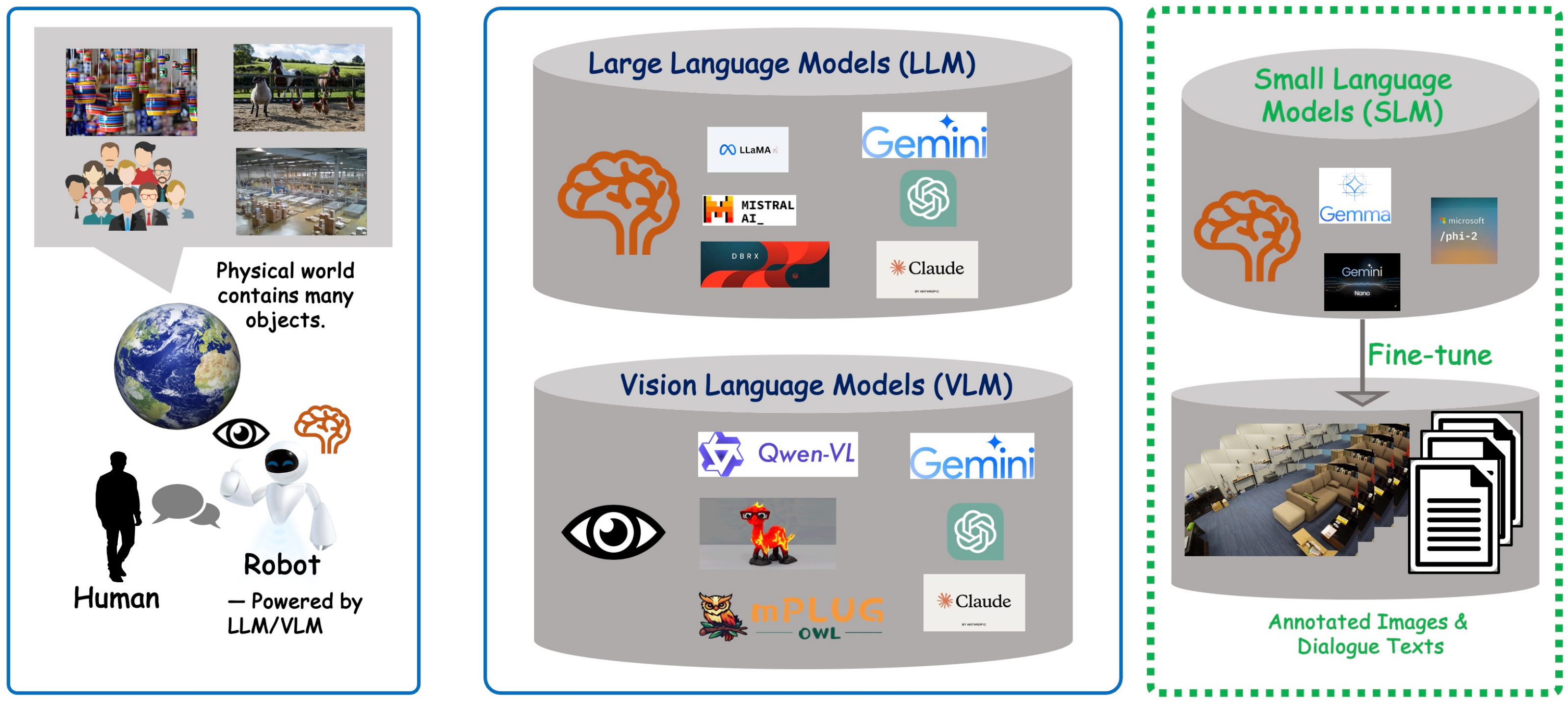}
  \caption{In HRI, robots acquire new abilities when roboticists incorporate  LLM/VLM into robots' \emph{dialogue and perception} modules. LLM lead to a better \emph{understanding} of dialogue context, world knowledge, and human utterances. Moreover, VLM enables robots to \emph{visually perceive} the gaze, emotion, and movements of humans and objects surrounding them. LLM examples include Llama, Mixtral, Gemini, GPT-4, and Claude. Similarly, VLM examples include Qwen-VL, LlaVa, mPLUG-Owl, Gemini, GPT-4V, and Claude. Unlike LLM, small language models (SLM), i.e., Gemma, Phi-2, and Gemini Nano, are better suited for robots due to fewer parameters.}
  \label{Figure:RobotLLMVLM}
\end{figure}

The fast-paced development of natural language processing (NLP) through LLM \cite{wan2024efficient, 10.1145/3641289, zhao2023surveylargelanguagemodels} promises to catalyze a paradigm shift in robotics research, for example, in how SARs are deployed. In addition, VLM\footnote{The jargon vision language models (VLM) and large multimodal models (LMM) are often used interchangeably in AI literature to refer to models that can process text+image or text+video data. However, LMM explicitly refers to models that process \textit{multiple modalities} not limited to text, audio, sound, proteins, graphs, or gene sequences.}  offer immense potential to enhance visual perception during human-robot interaction, leading to better situational awareness (see Figure \ref{Figure:RobotLLMVLM}). 
Several works have extensively studied the potential of LLM and VLM in robotics \cite{zeng2023largelanguagemodelsrobotics, wang2024largelanguagemodelsrobotics, hu2023generalpurposerobotsfoundationmodels, firoozi2023foundationmodelsroboticsapplications, kawaharazuka2024realworldrobotapplicationsfoundation}. 
However, we emphasize that before such development is widely adapted, it is crucial to understand the main applications of SARs and then identify key areas where LLM and VLM could be most helpful. Consequently, we conducted a meta-study of more than 250 HRI papers only from the \textit{Proceedings of the ACM/IEEE International Conference on Human-Robot Interaction}  spanning \textit{four years} from \textit{2020 to 2023}. We seek to answer these research questions: \textbf{RQ1:} What robots are most studied in HRI literature? \textbf{RQ2:} What applications do the robots serve? \textbf{RQ3:} What vital human values must be preserved when introducing LLM and VLM in SARs? \textbf{RQ4:} Can LLM components replace traditional language-processing robot components? \textbf{RQ5:} What are the benefits of integrating LLM into robot designs. Lastly, \textbf{RQ6:} what are the associated risks? By successfully answering the questions RQ1-RQ6 stated above, we aim to reach a common goal: \textbf{how to safely and responsibly adopt LLM and VLM to develop social robots.} 
\begin{figure}[!t]
  \centering
  \includegraphics[width=0.85\linewidth]{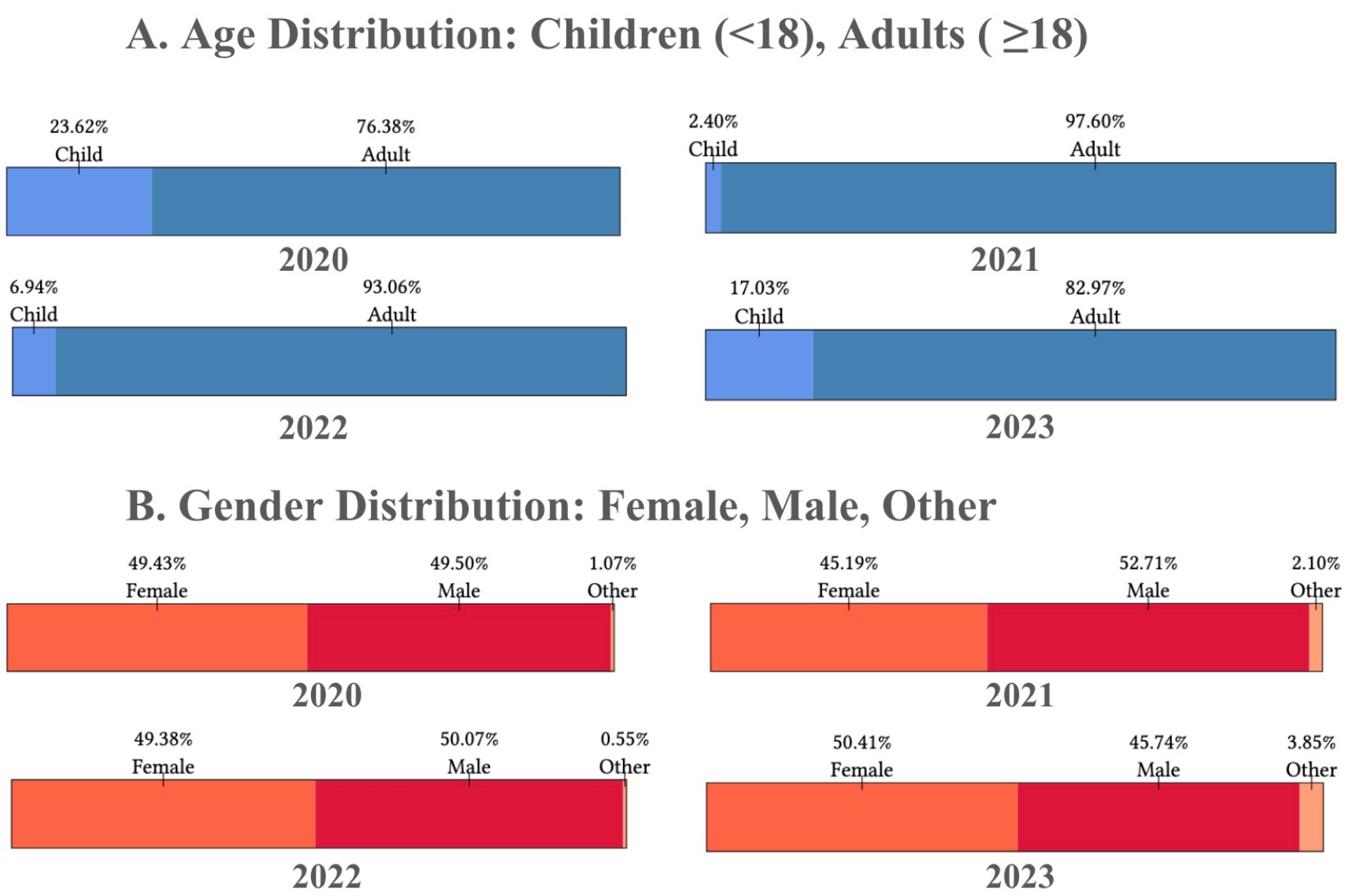}
  \caption{HRI papers in our meta-study included participants. We summarize the age and gender distribution of participant information from the papers. In the age distribution, we define children as those under 18 years old, while adults are 18 years and above. For gender distribution, we summarized female, male, and other.} 
  \label{Figure:Demographics}
\end{figure}
\begin{figure*}[t]
    \centering
    \resizebox{0.995\textwidth}{!}{
        \begin{tikzpicture}
            \definecolor{rangeColor}{RGB}{202,94,84}
            
            \fill[rangeColor] (0,0) rectangle (0.801,0.25);  %
            \fill[rangeColor!80] (0.801,0) rectangle (2.07,0.25);  %
            \fill[rangeColor!60] (2.07,0) rectangle (3.156,0.25);  %
            \fill[rangeColor!40] (3.156,0) rectangle (3.957,0.25);  %
            \fill[rangeColor!20] (3.957,0) rectangle (4.85,0.25);  %
            \fill[rangeColor!50] (4.85,0) rectangle (5.134,0.25);  %
            \fill[rangeColor!30] (5.134,0) rectangle (5.742,0.25);  %
            \fill[rangeColor!10] (5.742,0) rectangle (6.2136,0.25);  %
            \fill[rangeColor!70] (6.2136,0) rectangle (6.402,0.25);  %
            \fill[rangeColor!90] (6.402,0) rectangle (6.6378,0.25);  %
            \fill[rangeColor] (6.6378,0) rectangle (9,0.25);  %

            \draw[black] (0,0) rectangle (9,0.25);

            \foreach \position/\age/\percentage in {
                0.4005/0-9/8.9\%,
                1.4355/10-19/14.14\%,
                2.613/20-29/12.04\%,
                3.5565/30-39/8.9\%,
                4.4035/40-49/9.95\%,
                4.992/50-59/3.14\%,
                5.438/60-69/6.81\%,
                5.9778/70-79/5.24\%,
                6.3078/80-89/2.09\%, 
                6.52/90-99/2.62\%,
                7.8189/>100/26.18\%
            } {
                \node at (\position,0.35) [font=\tiny, rotate=45] {\age};
                \node at (\position,0.6) [font=\tiny, rotate=45] {\percentage};
                \draw[-] (\position,0.4) -- (\position,0.35);
            }
        \end{tikzpicture}
    }
    \caption{The papers in our meta-study often included participants. We summarize the distribution of the size of participant groups inside the HRI papers. Many studies included more than 100 participants.}
    \label{fig:AgeRange}
\end{figure*}

Despite the timely need to gather deep insights into LLM in social robotics, the potential and pitfalls of LLM in HRI are still underexplored. 
Among the earliest works to investigate LLM for HRI, the study conducted by \citet{ZHANG2023100131} is closest to ours, with notable differences: (i) our work focuses only on \textit{HRI conference} papers, (ii) we concentrate our efforts on \textit{conversational HRI}, and (iii) our study mainly dwells on the applications of social robots to gather insights about LLM usage in HRI. 
To collect the insights, our meta-study takes a systematic approach consisting of three phases. 
\\ \textbf{Phase I.} We gather overarching demographic insights about the age, gender, and number of participants found in HRI studies before thoroughly delving into the details of social robots typical of HRI studies, RQ1, and the applications of aforementioned robots, RQ2, in Section~\ref{sec:Applications}.  
\\ \textbf{Phase II.} We analyze human values, RQ3, often studied in HRI research, in Section~\ref{sec:HRI Key Societal Issues}. 
We observed that the \texttt{HRI conference} emphasizes key aspects of \textit{human-human} interactions, which are necessary to maintain harmony in society. These aspects, \textit{aka human values} include trust, ethics, teamwork, apology, safety, personality, politeness, inter alia. One of the objectives is to extend these human values to interactions between humans and robots. As a result, common themes at HRI conferences include: \textit{inclusive design and accessibility}, \textit{human-robot communication}, \textit{human-robot collaboration}, \textit{learning with robots}, \textit{human perception of robots}, \textit{robots for health and well-being}, among others. These themes serve to guide roboticists in developing robots that can correctly perceive societal norms and practices across cultures. 
\\ \textbf{Phase III.} After acquiring insights about RQ1, RQ2 \& RQ3, we sought to investigate the potential of LLM in HRI. However, it is important to understand conventional robot designs before investigating the potential and pitfalls of LLM in HRI. Hence, in Section \ref{section:LLMandVLMinRobotDesign}, we present an in-depth analysis of vital language components of a SAR, and we discuss what components the LLM or VLM could replace, RQ4. Thereafter, we present the potential benefits of LLM in HRI applications, RQ5, in Section \ref{sec:LLM_in_HRI_Opportunities}. Lastly, in Section \ref{sec:LLM_in_HRI_Risks}, we discuss the downsides of LLM if deployed in HRI, RQ6.

Throughout this meta-study, we envision the adoption of LLM in HRI through the lenses of specific applications of SARs, such as education, healthcare, entertainment, and the hospitality or services industries. 
\section{Demographic Data}
Human participants are essential to HRI work.\footnote{Several HRI papers do not fully disclose information about the participants. Hence, we present details only from papers that provided this information.} Our analysis revealed that most studies recruit adults, i.e., 18 years and above, and children are less involved in HRI studies. However, HRI researchers recruit a fairly balanced number of women and men in the studies, see Figure \ref{Figure:Demographics}. Moreover, on top of \textit{female} and \textit{male} gender definitions, other gender lingo in HRI studies include \textit{gender fluid, diverse, agender, nonbinary, prefer not to say}.   
The number of participants in HRI studies varies greatly, and 26\% of the studies reported more than 100 participants. 
\section{Applications of Social Robots}
\label{sec:Applications}
In this section, we describe the major robots studied in HRI, and their applications. 
\subsection{Robots in HRI Research}
\label{sec:Robots in HRI Research}
\begin{figure*}[!t]
  \centering
  \includegraphics[width=0.95\linewidth]{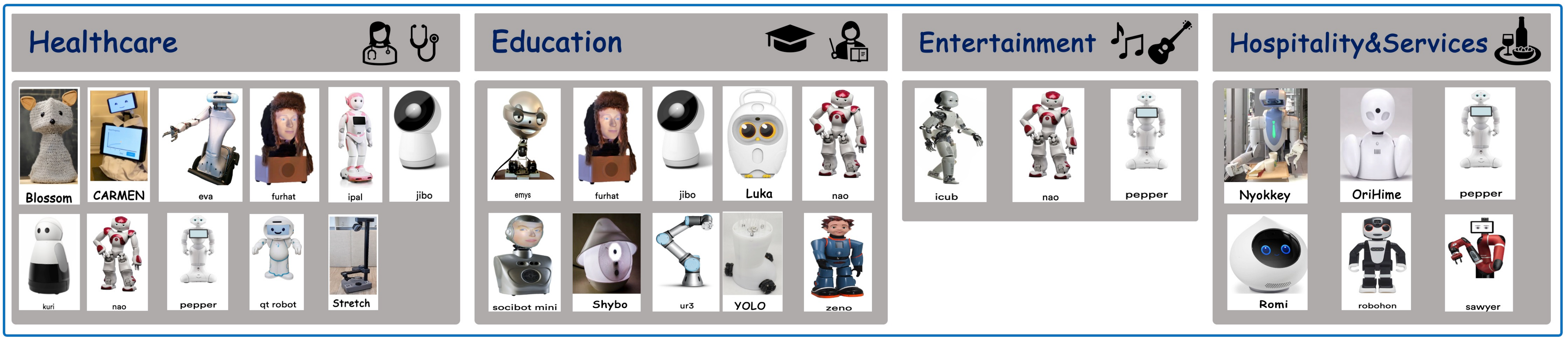}
  \caption{Examples of socially assistive robots (SARs) encountered in our study. The robots are deployed in healthcare, education, entertainment, and hospitality\&services applications. (Most images are taken from the \textit{ABOT Database}). } 
  \label{Figure:RobotsOverview}
\end{figure*}

\textbf{RQ1. What robots are studied in HRI research?}
Our analysis reveals that a diverse range of robots are employed in HRI (see Figures \ref{Figure:RobotsOverview}, \ref{fig:Robots_in_HRI_research}), with Nao and Pepper appearing regularly in  HRI works, though the HRI field is not homogeneous and other robots include Robovie, Kuri, Fetch, Sawyer, iCub, Furhat, UR, and Jibo. (All robots are shown in Appendix \ref{Appendix:ExemplarRobotsinHRIResearch}, Table \ref{Table:Common_Social_Robots_HRI}.) The robots serve different roles, including rehabilitation robots, shopping agents, and tutors.
\begin{figure}[!t]
\raggedright
\begin{tikzpicture}
\begin{axis}[%
width=0.95\linewidth, height=1.8in,
xbar, bar width=4.2pt,
xmin=0, xmax=110,
symbolic y coords={Nao, Pepper, Jibo, UR, Furhat, iCub, Sawyer, Fetch, Kuri, Robovie, Others},
ytick=data,
yticklabels={Nao, Pepper, Jibo, UR, Furhat, iCub, Sawyer, Fetch, Kuri, Robovie, Others},
y tick label style={font=\tiny, align=right, text width=1.2cm},
xtick={0,20,40,60,80,100},
xmajorgrids,
axis line style={lightgray},
major tick style={draw=none},
nodes near coords,
point meta=explicit symbolic,
node near coords style={font=\tiny, right=1em, pin={[pin distance=1em]180:}}
]
\addplot [fill=brown!60, draw=none] coordinates {
                (19,Nao) [19 (11\%)]  
                (15,Pepper) [15 (8\%)] 
                (10,Jibo) [10 (6\%)] 
                (9,UR) [9 (5\%)]
                (8,Furhat) [8 (5\%)]
                (6,iCub) [6 (3\%)]
                (6,Sawyer) [6 (3\%)]
                (5,Fetch) [5 (3\%)]
                (5,Kuri) [5 (3\%)]
                (5,Robovie) [5 (3\%)]
                (90,Others) [90 (50\%)]
                };
\draw [line width=1.5pt] (current axis.south west) -- (current axis.north west);
\end{axis}
\end{tikzpicture}
\caption{Robots in HRI research. The horizontal axis indicates the number of times each robot appeared during our study and the percentage inside parenthesis.}
\label{fig:Robots_in_HRI_research}
\end{figure}
\subsection{Applications}
\textbf{RQ2. What applications are social robots mainly used for?}
Next, we discuss examples of SARs\footnote{The ABOT Database at \url{https://www.abotdatabase.info/collection} is a good collection of robots.} and their applications, across five domains, notably in \textbf{healthcare, education, entertainment, hospitality \& services,} and finally, \textbf{tele-operations \& telepresence}. (See Figure \ref{Figure:RobotsOverview}).
These are the examples of robots in each domain. 
(i) Healthcare: \textit{Kuri, CARMEN, EMAR, Pepper, Eva, Nao, Jibo, Furhat, iPal, QT robot, Stretch, and Blossom} (Section~\ref{sec:Healthcare}).
(ii) Educational studies: \textit{Zeno R25, SocibotMini, Nao, YOLO, Luka, EMYS, UR5e, Shybo, Jibo, and Furhat} (Section~\ref{sec:Education}).
(ii) Entertainment: \textit{iCub, Nao, and Pepper} in  (Section~\ref{sec:Entertainment}).
(iv) Hospitality \& services: \textit{Lovot, Pepper, OriHime, OriHimeD, Sawyer, Romi, Charlie, RoBoHoN, Nyokkey, and Sota} (Section~\ref{sec:Hospitality}).
(v) Telepresence: \textit{Double and Beam} (Section~\ref{sec:TeleoperationsTelepresence}).
We have included examples of robots per application category in Appendix \ref{Appendix:ApplicationsOfRobots}, Table \ref{Table:ApplicationsofAssistiveRobots}).
\subsubsection{Healthcare}
\label{sec:Healthcare}
SARs have been widely adopted to study illnesses such as stroke, depression, autism; investigate robot-assisted feeding; and to deliver therapy sessions, among other use-cases.
One line of work studied the potential of robots to rehabilitate \cite{10.1145/3319502.3374836, 10.1145/3568162.3576993} or deliver cognitive training \cite{10.5555/3523760.3523771} for people with mild cognitive impairment (MCI).  The authors deployed \texttt{Kuri, CARMEN and EMAR} robots. 
Stroke is another use-case in HRI studies. \cite{10.1145/3319502.3374797} developed an interactive game to rehabilitate stroke patients using \texttt{Pepper}, while \cite{10.1145/3319502.3374775} experimented with remote diagnosis of stroke patients. Moverover, \cite{10.1145/3319502.3374840} studied the impact of \texttt{Eva} on reducing behavioral and psychological symptoms of dementia (BPSD) for people with dementia. 
Researchers have also investigated the use of SARs for children's healthcare where autism \cite{10.5555/3523760.3523766, 10.5555/3523760.3523770, 10.1145/3568162.3576963}  is a major topic, while rehabilitation \cite{10.1145/3319502.3374826}, and anxiety/depression \cite{10.1145/3568162.3576960} among children were also studied. \texttt{Jibo, Nao, Furhat, iPal} robots featured in these studies. 
Researchers have studied robot-assisted feeding \cite{10.1145/3319502.3374818} and applied robots to feed people with mobility impairments. 
\citet{10.1145/3434073.3444671} used \texttt{Jibo} to deliver daily positive psychology sessions to undergraduate students.
Another interesting line of work studied the use of robots to understand the mental well-being of humans \cite{10.1145/3568162.3578625}, automated delirium detection \cite{10.1145/3568162.3576971}, and to deliver positive psychology exercise \cite{10.1145/3568162.3577003}. These studies included \texttt{Jibo, Misty II, QT} robots. 
Lastly, researchers have studied the effectiveness of mobile telemanipulator robots (MTRs) in hospital emergency departments \cite{10.1145/3568162.3576994}, to support multitasking in environments with frequent task switching.
\subsubsection{Education} %
\label{sec:Education}
In the education sector, teachers are using robots to: 
(i) teach handwriting e.g., for the Kazakh alphabet \cite{10.1145/3319502.3374813} using \texttt{Nao}, 
(ii) teach vocabulary using tutee robots \cite{10.1145/3319502.3374822}, and to encourage good word retention \cite{10.1145/3568162.3577004} via  word-learning card-games based on \texttt{Furhat}, 
(iii) teach a second language to students \cite{10.1145/3434073.3444670},
(iv) teach musical instruments \cite{10.1145/3319502.3374787} using \texttt{SocibotMini},
(v) increase student engagement \cite{10.1145/3319502.3374815}  using \texttt{Nao's} gestures, and increase children's creativity \cite{10.1145/3319502.3374817, 10.5555/3523760.3523774} using \texttt{YOLO, EMYS} respectively, 
(vi) stimulate learning and early development among toddlers \cite{10.1145/3319502.3374792}, and to teach toddlers to read \cite{10.5555/3523760.3523768} using \texttt{Luka}. 
A social robot companion based on \texttt{Jibo} was used to motivate children to read storybooks \cite{10.1145/3568162.3576968},
(vii) promote critical thinking among school children \cite{10.5555/3523760.3523837} using \texttt{Shybo},
(viii) leverage robot tutors and teach maths \cite{10.1145/3568162.3576957} using  \texttt{Nao} or teach a language such as Japanese \cite{10.1145/3568162.3578633} using \texttt{Furhat}.
Lastly, researchers have also studied the interactive behavior between students and robots \citet{10.1145/3319502.3374803} by deploying \texttt{Zeno R25}.
\subsubsection{Entertainment}
\label{sec:Entertainment}
\citet{10.1145/3319502.3374780} studied the humorous part of robots by enabling \texttt{NAO} to tell jokes to an audience. \citet{10.1145/3319502.3374809} programmed robots to take photos in portrait mode. Whereas in \cite{10.1145/3319502.3374837}, a wearable robotic device that creates a cyborg character was involved in a dance performance. Meanwhile, \citet{10.1145/3434073.3444682} developed a framework that enabled \texttt{iCub} to autonomously lead an entertaining and effective human-and-robot interaction, based on the real-time reading of a biometric feature from the players. Lastly, \citet{10.5555/3523760.3523834} proposed a new understanding of improvisation based on rules that shape robot movement and behavior, leading to increased engagement and responsiveness in a dance performance.
\subsubsection{Hospitality and Services}
\label{sec:Hospitality}
In this domain,  researchers have investigated the use of robots in restaurants, robot cafes, bakery stores, clinics, convenient stores, and food delivery.
\citet{10.5555/3523760.3523831} trained \texttt{Pepper} to behave appropriately as a \textit{waiter} in a restaurant and to respond to customer requests. 
In addition, \citet{10.5555/3523760.3523829} investigated the influence of several forms of social presence of teleoperated robots on customer behavior. They discovered that the robot, which exhibited a moderate presence of the operator (costume form), achieved the best overall performance. 
\citet{10.1145/3568162.3576967} studied environments where several robots were deployed as service robots in \textit{robot cafes}. The robots included \texttt{Lovot, Pepper, OriHime, OriHimeD, Sawyer, Romi, Charlie, RoBoHoN, and Nyokkey}.
Moreover, \citet{10.1145/3568162.3577005} found that deploying service robots, named \texttt{Sota}, in pairs increased sales at a bakery store.  
\citet{10.1145/3319502.3374795} developed their own robot to study the impact of \textit{co-embodiment} and \textit{re-embodiment} in the services domain, such as Quick Care Clinic, Canton Department Store, and Homestead Inn. 
Lastly, \citet{10.1145/3568162.3576984} deployed a delivery robot to dispatch food and convenience store products to customers at various locations. They further investigated the impact of groups towards acceptance and trust of robots, revealing that groups were more hesitant to accept robots yet individual users trusted them. 
\subsubsection{Tele-operations and Telepresence}
\label{sec:TeleoperationsTelepresence}
\citet{10.1145/3568162.3576961} studied the lived experience of participating in hybrid spaces through a telepresence robot. The robots used in this study were \texttt{Double} and the \texttt{Beam}.
\section{Key Societal Norms and Human Traits in HRI}
\label{sec:HRI Key Societal Issues}
\subsection{Overview}
We have presented examples of social robots that are studied in HRI research (RQ1) and also discussed the applications of social robots (RQ2) in Section \ref{sec:Applications}. 
Our discussion shows that HRI studies focus on developing robots to interact with humans. 
To achieve successful interactions, under a wide range of applications and robots, it is vital to preserve \textit{social norms} and \textit{human traits} that humans hold so dearly. These norms are essential to keep harmony among humans, and it is beneficial to replicate these norms when designing robots intended to communicate with humans. LLMs provide a promising avenue to incorporate these \textit{human values} into social robots. 
\\ \textbf{RQ3. What vital human values must be preserved when introducing LLM/VLM in SARs?}
Examples of human values include \textit{trust, politeness, personality, etc.} 
LLM provide roboticists with plentiful options to introduce human values into robot design. 
\textit{First,} LLM can be fine-tuned to embody \textbf{trust} behavior before deployment in a robot. In addition, prompting provides another dimension for roboticists to guide LLM in eliciting behavioral types necessary to establish trust between humans and robots during interaction. Establishing trust includes trust repair strategies, such as apology, denial, explanation, and promise \cite{deVisser2018FromT}.   
\textit{Second,} another trait highly cherished by humans is \textbf{politeness} \cite{10.1145/3568162.3576959}. LLM can be fine-tuned or prompted to generate polite responses during SAR development.  SARs need to be flexible to accommodate several \textbf{personalities}. LLM can be beneficial by providing the appropriate word choice, speed, frequency of gestures, and the robot's form necessary to influence the perceived personality of the robot by humans \cite{10.1145/3424153, 10.1145/3439795, 10.1145/3424153, DBLP:journals/jmui/BevacquaSHP12, 10.5555/2615731.2617415, Broadbent2013RobotsWD, 10.1145/3568162.3577003, gao-etal-2023-peacok}. \textit{Third,} SARs need to excel at \textbf{gender} identification to encourage smooth interactions with humans. 
\subsection{Trust}
According to the Cambridge dictionary, to trust is \textit{``to believe that someone is good and honest and will not harm you, or that something is safe and reliable''}. Hence in HRI terms too, humans expect the robots they interact with to be ``safe'' and ``reliable''. Trust is heavily studied in HRI and without a clear definition for trust in the HRI realm, trust is often seen as ``a multidimensional psychological attitude involving beliefs
and expectations about the trustee’s trustworthiness derived from experience and interactions with the trustee in situations involving uncertainty and risk''~\cite{f5dfd909-509b-3958-bbd0-45346f72ee41, Lewis2018}. 
It is important to note that trust evolves with time, and trust is characterised by three phases~\cite{Lewis2018, 10.1145/3472224}; trust formation, trust dissolution and trust restoration. When an interaction starts, user-trust is built upon the robot’s appearance, context information, and the person’s prior experience with robots. Then trust dissolution occurs during the interaction when users lower their trust in the robot due to a trust violation, e.g., robot error. Lastly, trust restoration occurs when trust is repaired and therefore user-trust stops decreasing after a trust violation.
Given the significance of trust in HRI, researchers have devised several strategies to repair trust of robots by humans during an interaction. The \textit{trust repair strategies} are; apology, denial, explanation, and promise. The effectiveness of these strategies has been studied~\cite{deVisser2018FromT}. In this work, we emphasize that these four trust-repair strategies should be incorporated into the LLM e.g., via prompting or fine-tuning on datasets containing such data instances, because the quality of continued HRI heavily depends on how the robot repairs lost user-trust due to trust violations~\cite{10.5555/3378680.3378691}.
\subsection{Personality}
Humans exhibit heterogeneous personalities, and this behavior needs to be passed on to social robots, too. 
Robot personality has been achieved by adjusting behavioral parameters such as word choice~\cite{10.1145/3424153, 10.1145/3439795}, speed~\cite{10.1145/3424153} and frequency of gestures~\cite{DBLP:journals/jmui/BevacquaSHP12, 10.5555/2615731.2617415}. Moreover, the robot's form \cite{Broadbent2013RobotsWD} can further influence the perceived personality of the robot by users.
\citet{10.1145/3568162.3577003} developed a robotic coach to conduct positive psychology exercises to promote the mental well-being of employees at a company. 
Furthermore, LLM can be calibrated to exhibit or elicit several kinds of personalities (e.g., the personality of a mental health coach)  suitable for a user application. For example, \citet{gao-etal-2023-peacok} prompts the InstructGPT-3 LLM to generate new attributes for different kinds of persona, including actor, singer, politician, etc.
Politeness is another vital human trait. \citet{10.1145/3568162.3576959} studied politeness using \texttt{Pepper}, and their goal was to examine the influence of wakewords on politeness directed by either humans or robots. Politeness should be incorporated into LLM e.g., via prompting, to encourage polite responses.
Therefore, roboticists can leverage LLM prompting to adjust SAR personality during HRI.

\subsection{Gender} Given the diversity of culture and the sensitivity towards topics such as \textit{gender} across cultures, social robots need to be designed considering this factor. LLM can be vital in being trained or prompted to respond appropriately to a user's gender identity. It is equally important to deploy affirmative action strategies to minimize stereotypical thinking and biased perceptions of certain genders \cite{10.1145/3568162.3576977, 10.1016/j.chb.2014.05.014}.

\section{LLM and VLM in Robot Design}
\label{section:LLMandVLMinRobotDesign}
After discussing SARs (RQ1), their applications (RQ2), and human traits relevant to HRI (RQ3), we turn our attention to LLM deployment in SAR design and development (RQ4). To successfully integrate LLM into SARs, it is essential to understand how conventional SARs are built and identify areas where LLM could be most relevant. 
Primary natural language processing (NLP)  modules inside robots include automatic speech recognition (ASR), user-intent classification (IC), dialogue management (DM), and text-to-speech synthesis (TTS). In addition, SARs contain several sensor modules for perception. Below, we describe conventional designs for social robots in Section~\ref{sec:Conventional Designs}. Then, we tackle RQ4 by describing ways to leverage LLM in robot design, in Section~\ref{sec:LLM-based Design}. 
\subsection{Conventional Robot Architecture}
\label{sec:Conventional Designs}
We present four SAR examples: Nao, Sophia, ARI, and Butsukusa, among the many SARs in our study.
\subsubsection{Nao} %
This is a humanoid NAO robot manufactured by SoftBank Robotics. It is a widely-used social robot in HRI research (Section~\ref{sec:Robots in HRI Research}), offering a flexible programming environment for researchers. It has basic modules, such as, built-in speech recognition, face recognition, display of gestures and body postures, and a multilingual text-to-speech engine that enables it to function more naturally. 
\subsubsection{Sophia}
\label{subsubsection:Sophia}
\begin{figure}[h]
  \centering
  \includegraphics[width=0.85\columnwidth]{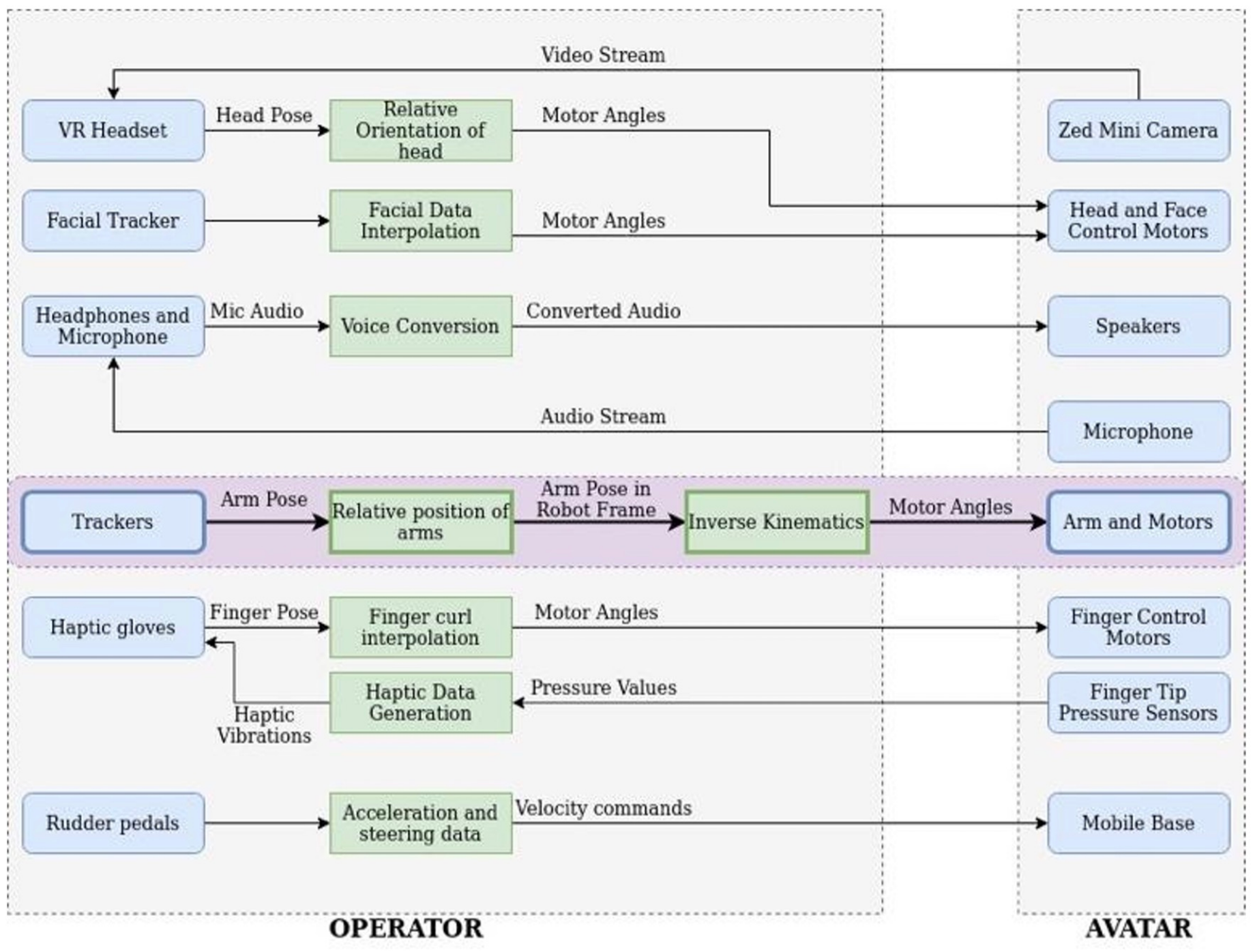}
  \caption{Key components of the Sophia robot.} 
  \label{Figure:sophiaRobot}
\end{figure}

As shown in Figure \ref{Figure:sophiaRobot}, the architecture of Sophia \cite{hanson2022openarmsopensourcearms} comprises several modules. The robot acquires \textit{situational understanding} through multimodal learning of vision, speech, grasping, manipulation, locomotion, and social interactions with human users. For instance, audio and vision perception happens through headphones and cameras. A stereo camera records a video stream, while a microphone array records the audio stream. A separate facial tracking module tracks lips and mouth movements. Respective modules then process these modalities. In addition, the ensemble of verbal and non-verbal dialogue models is managed by neural frameworks combined with a rules-based controller. 
\subsubsection{ARI}
\citet{10.1145/3568294.3580041} developed a social robot, ARI, entirely based on NLP, and the robot's two main components are \textit{vosk package} for ASR in multiple languages, and \textit{rasa} to manage dialogues. Consequently, the robot is characteristic of an NLP pipeline fully compatible with ROS and ROS4HRI standards. It features a dedicated language center for internationalization, handling textual translations, and model swapping for ASR/TTS/DM. The pipeline introduced the concept of user intents, distinct from chatbot intents, which encapsulate user-initiated commands, allowing separation between user intention recognition nodes and the robot's application controller. This design fosters code sharing and enhances controller reusability across robot platforms. Lastly, the pipeline is integrated with a knowledge base and reasoning framework to facilitate dialogue management with real-time events the robot senses.
\subsubsection{Butsukusa}
\label{subsubsection: Butsukusa}
\begin{figure}[ht]
  \centering
  \includegraphics[width=0.95\columnwidth]{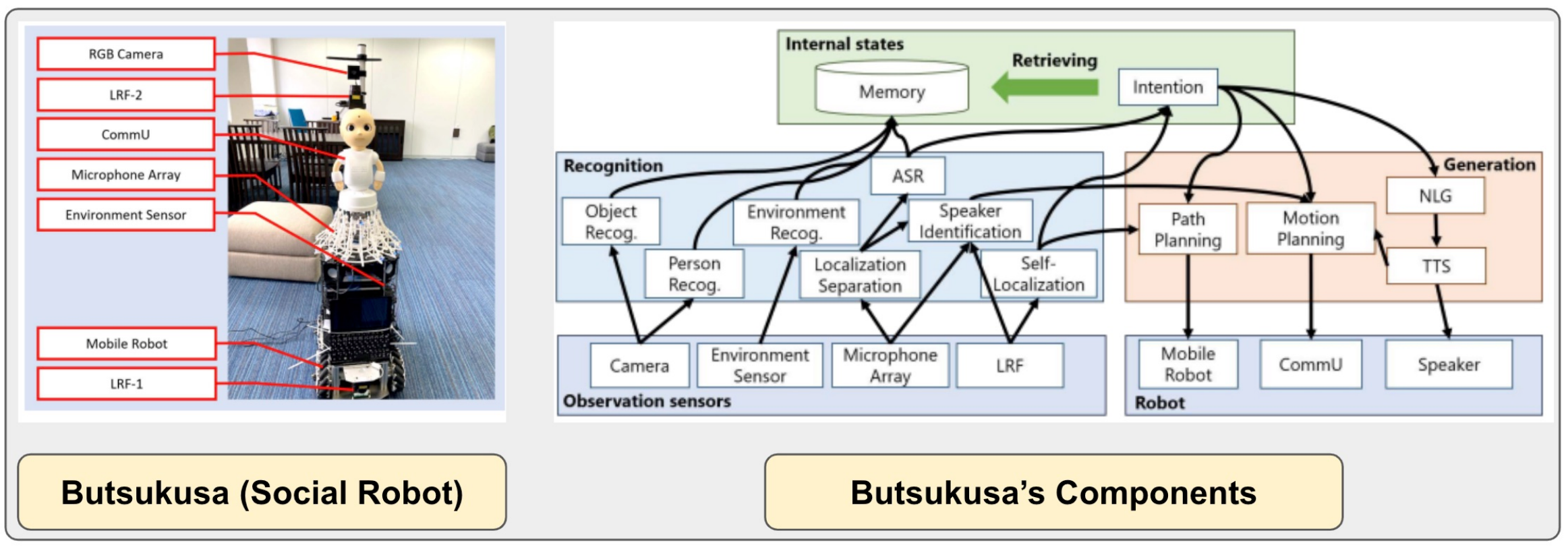}
  \caption{Components of Butsukusa: information processing, recognition, internal states, generation, robot.} 
  \label{Figure: Butsukusa Design}
\end{figure}

The components of Butsukusa~\cite{10.5555/3523760.3523946} are shown in Figure~\ref{Figure: Butsukusa Design}, and they include (1) \textit{recognition module} which performs object recognition with \textit{mask R-CNN}; person recognition with a \textit{CNN}-based feature extractor; environment recognition with a sensor; sound localization, speech separation, speaker identification; automatic speech recognition (ASR); and self-localization. The observations are stored in memory. (2) The \textit{internal states module} consists of memory in which results for persons, objects, ASR, and self-localization are stored in the form of an SQL database. Moreover, Intention defines the choice of next action taken by the robot. (3) The \textit{generation module} receives the intention and then generates the next action of the robot via path planning, natural language generation, speech synthesis and motion planning. Whereas this robot could be easy to maintain and debug due to ease of localizing robot failures, the robot architecture is complex. LLM and VLM can be used to reduce this complexity.
\subsection{LLM-based Robot Architecture}
\label{sec:LLM-based Design}
A notable shift in SAR design towards LLM-powered architectures is ongoing, promising enhanced interaction capabilities through improved speech recognition, intent detection, dialogue management, and multi-modal preception. Consequently, ASR, IC, DM, TTS, and multimodal perception components are candidates for replacement by LLM and VLM. 
\\ \textbf{RQ4: Can LLM replace traditional robot components?}
From the robots presented above, there is opportunity to harness LLM to build robots, replacing conventionally used cloud services such as, Google Speech for ASR, and Rasa / Google DialogFlow for  dialogue management. 
The following discussion presents vital processes in social robots, and the role LLM or VLM can play to advance these processes. \\
\textbf{Automatic Speech Recognition (ASR).}
ASR converts spoken language into written text, and it is crucial for understanding user commands in spoken interactions. Common ASR tools include Google Cloud Speech-to-Text which offers powerful and accurate speech recognition; IBM Watson Speech to Text which converts audio and voice into written text; Microsoft Azure Speech which provides speech-to-text services for real-time and batch processing; and Mozilla DeepSpeech, an open-source ASR engine based on deep learning. Available LLM with multimodal and multilingual abilities are well-positioned to perfom ASR with higher ASR accuracy, support for multiple languages, and processing of both speech and textual data. \\
\textbf{User-Intent Classification (IC).}
Intent classification is the process of determining what the user wants to achieve with their input. Several methods have previously been used to detect user intent. NLP techniques like tokenization, lemmatization, and parsing for understanding user intent; machine learning classifiers, such as SVM, Naive Bayes, and Decision Trees, are used to categorize user input into specific intents; deep learning models like RNNs and Transformers are used for more complex intent detection tasks; and Rasa NLU, a natural language understanding solution for intent classification and entity extraction in conversational AI.  The impecable abilties of LLM to classify intent with high accuracy means that we can reduce the complexity of IC mdules by using LLM. \\
\textbf{ Dialogue Management (DM).}
Dialogue management is responsible for managing the conversation between the human and the robot, deciding what the robot should do or say next based on the user’s input. Examples include: rule-based systems which use predefined rules to manage dialogue flow; state machines which define possible states and transitions in a conversation; reinforcement learning which trains models to optimize dialogue flow based on rewards; and Rasa, an open-source machine learning framework for building contextual AI assistants and chatbots. LLM are well-suited for this task. \\
\textbf{Text-to-Speech Synthesis (TTS).}
TTS converts written text into spoken words, enabling the robot to communicate audibly with users. TTS examples include Google Text-to-Speech, which converts text into human-like speech using deep learning; Amazon Polly, which turns text into lifelike speech using advanced deep learning technologies; IBM Watson Text Speech, which converts written text into natural-sounding audio; and Microsoft Azure Text Speech, which provides natural-sounding voices for rendering text into speech. Similar to ASR, multimodal LLM are getting better at processing speech and text data so TTS modules can be replaced by LLM.\\
\textbf{Object Recognition.} Available LLM and VLM can correctly identify objects from both textual and visual data and ground the location of objects occurring in the textual utterances to the objects' locations inside a visual scene. \\
\textbf{Face, Gaze, Emotion, and Pose Recognition.}
Due to improved multimodal learning methods and VLM, previously independent tasks such as the recognition of faces, gaze, emotions, and human body poses  can be performed by VLM.\\ 
\\ \textit{In summary, previously independent components described above can be replaced with LLM or VLM, which have shown impeccable abilities for multimodality, tracking dialogue histories, classifying intent, recognizing faces, gaze, emotions, and human body poses, detecting and grounding objects, etc}. Such integrated designs will reduce the complex architectures of new SARs.
\section{LLM Benefits in HRI}
\label{sec:LLM_in_HRI_Opportunities}
\textbf{RQ5. What are the benefits of integrating LLM into robot designs, and what are the associated risks?}
After discussing the components of SAR design in which LLM can be deployed, we present the advantages of LLM in robots.
\subsection{Wealth of Knowledge} The wealth of knowledge contained inside LLM \cite{petroni2019languagemodelsknowledgebases} translates into improved abilities for robots to comprehend diverse topics during human-robot dialogue. For instance, the LLM knowledge about autism or cognitive impairment can adapt a robot's dialogue responses to patients with such illnesses.  
\subsection{Reasoning in LLM}
\textbf{Reasoning over Facts.} Integrating knowledge graphs and advanced prompting techniques with LLM \cite{agrawal-etal-2024-knowledge, Pan_2024} can enhance a robot's ability to use factual reasoning, mitigating hallucination to improve the accuracy of human-robot interactions.
\\ \textbf{Complex Reasoning.} In RT-2~\cite{rt22023arxiv}, researchers posit that \textit{``incorporating chain-of-thought reasoning allows robots to perform multi-stage semantic reasoning like deciding which object could be used as an improvised hammer (a rock), or which type of drink is best for a tired person (an energy drink)''}. Advanced LLM has shown impeccable abilities for complex reasoning, crucial for robots.
\subsection{Planning}
\citet{knowno2023} demonstrated LLM planning ability by introducing an LLM-based framework, \textit{KNOWNO}, which combines few-shot prompting and Bayesian inference to align the uncertainty of LLM-based planners with the true uncertainty of the task environment. Then, robots performed complex multi-step planning tasks with statistical guarantees of task completion with minimal human help.  The framework outperformed modern baselines on efficiency and autonomy when tested on mobile manipulation, navigation, and question-answering. Such LLM-based planning improves robot performance. 
\subsection{Personalization}
By creating multiple \textit{persona}, LLM facilitates the personalization of robots to meet user-specific needs.
\citet{10.5555/3523760.3523775} showed that users kept engaged with the social robot if the robot remembered user names, interests, and opinions. 
LLM can fulfill this requirement, making social robots more fun to interact with. 
Moreover, \citet{10.1145/3568162.3578624} emphasized that robot design for persons living with dementia (PwD) should enable slow communication between the robot and the PwD. That way, PwD feel respected. LLM can be prompted to deliver dialogue responses at a slow pace when interacting with PwD, and to be selective in what information it reminds the PwD. Hence, LLM foster increased personalization of robot dialogue.
\subsection{Multimodality}
\label{Multimodality}
Large multimodal models (LMM) and VLM are vital to enable SARs to process multiple data modalities from cameras, microphones and other sensors mounted onto the robot.
\citet{10.5555/3523760.3523776} pointed out that interactions between user and robot include several sensory signals/feedback: tactile, auditory, and visual. The increasingly multimodal nature of LLM makes them suitable for the integration of these signals. 
Similarly, \citet{brohan2022rt1} introduced the \textit{Robotics Transformer} which comprises image tokenization, action tokenization, and token compression, resulting in a model that processes both text instructions and images, and then encodes them as tokens before compressing them with a TokenLearner. After this, tokens are fed to the Transformer. The Transformer then outputs the action-tokens. 
More recently, \citet{rt22023arxiv} introduced RT-2, a vision-language-action (VLA) model obtained by fine-tuning PaLM-E~\cite{driess2023palme} and PaLM-X, in which robot actions are represented as tokens, in addition to vision-language tokens. The resultant model, i.e. RT-2 achieved great results on human recognition, reasoning, and symbol understanding. RT-2 opened new possibilities for end-to-end robotic control.
Moreover, \citet{karamcheti2023voltron} introduced a new framework, \textit{Voltron}, for language-driven representation learning from large video datasets of humans performing everyday tasks. The framework balances conditioning and generation to shape the balance of low and high-level features captured. 
The availability of these frameworks enables integration of features from images, text, videos, audio, etc., to develop dialogue robots.
\citet{10.1145/3568162.3576958} developed a robot presenter that integrates head movements, speech, facial expressions, body pose and gaze to respond to the audience. Their \texttt{Furhat} robot utilizes a knowledge graph and GPT-3 \cite{NEURIPS2020_1457c0d6} to generate natural language explanations and adapt to user feedback.
\subsection{Speech and Language Components}
\label{Speech and Language Processing for Robot Dialogue Components}
LLM are essential to construct effective speech processing components, i.e., automatic speech recognition, speaker identification, and language processing components, i.e., sentiment analysis, semantic textual similarity and topic classification, as shown in \cite{10.5555/3523760.3523790}. 
In addition, \citet{10.1145/3568294.3580040} introduced one of the earliest adoptions of LLM in social robots by utilizing \textit{GPT-3 Davinci} to support the dialogue systems of both \texttt{Nao} and \texttt{Pepper} robots. By combining \textit{GPT-3 Davinci}, \textit{Google Cloud Speech-to-Text}, and \textit{NaoQi text-to-speech}, the authors transformed the English text-based interaction of GPT-3 into an open verbal dialogue with the robot.
\subsection{Synthetic Dataset Construction}
\label{Automate Dataset Construction}
LLM create large, high-quality synthetic datasets, crucial to robot development.
\citet{karamcheti2023voltron} used ChatGPT prompts to generate many language instructions for several actions. 
In addition, \citet{xiao2022robotic} deployed a fine-tuned VLM to construct a large dataset of language descriptions needed to train a robot policy.  This approach cost less time and money compared to human annotators.
\subsection{Reinforcement Learning with Human or Artificial Intelligence Feedback}
LLM techniques such as reinforcement learning from human feedback (RLHF)\cite{christiano2023deepreinforcementlearninghuman, ouyang2022traininglanguagemodelsfollow} and reinforcement learning from AI feedback (RLAIF) \cite{bai2022constitutionalaiharmlessnessai} help to align LLM behavior with human needs, which is crucial for SAR development where \textit{human trust} must be earned by the robot. 
For example, \citet{10.1145/3319502.3374795} trained a robot waiter in a restaurant application to adapt to customer requests. 
\subsection{Lifelong Learning and Embodied Agents}
LLM are vital to developing embodied agents.
\citet{wang2023voyager} introduced a LLM-powered embodied lifelong learning agent. This GPT-4 based agent, \textit{VOYAGER}, continuously explores the world, acquires highly sophisticated skills, and makes new discoveries consistently without human intervention. \textit{VOYAGER} does not require tuning of model parameters, contributing to a new frontier of prompting-based learning in embodied agents.
\subsection{Instruction Finetuning}
This method makes it possible to adapt LLM or VLM to a very specific task with minimal effort \cite{wei2022finetunedlanguagemodelszeroshot}. The only effort is to create an \textit{instruction-dataset} from existing datasets, and use the new data to fine-tune LLM to respond to instructions such as \textit{summarize this dialogue}, \textit{translate this text}, etc. %
\subsection{Faster Inference and LLM Optimization} 
LLM optimization methods, such as quantization, PEFT, knowledge distillation, neural architecture search, pruning, knowledge distillation, conditional computation, etc., promise to reduce LLM parameter count, deploy fewer attention network layers, and reduce memory requirements to adapt LLM to new tasks. Reduced LLM sizes make LLM suitable for deployment in social robots due to faster inference. %
\subsection{Robot Manipulation}
LLM continue to play a pivotal role in robotic manipulation. 
\citet{10.1145/3568162.3578623} achieved online robot manipulation by deploying the Distil-RoBERTa~\cite{Sanh2019DistilBERTAD} LLM to find the most similar training utterances to a given natural language correction, and utilizing the utterances to generate a corresponding latent action. The process enabled the system to quickly and accurately adapt to natural language corrections during execution. 
In addition, the authors used  GPT-3~\cite{NEURIPS2020_1457c0d6} prompts to output the degree of context-dependence required, and used it as a preprocessing step before training the system. The use of LLM eliminated the need to rely on heuristics or grammars. 
\subsection{Retrieval-Augmented Generation}
Retrieval augmented generation (RAG) \cite{lewis2021retrievalaugmentedgenerationknowledgeintensivenlp} in LLM is crucial in robot development. 
RAG-based LLM can access and manipulate external knowledge sources more effectively than other pre-trained LLM, making it a promising approach for human-robot dialogues tasks that require specialized domain knowledge. 
The RAG-based LLM  ability to hot-swap the retrieval index makes it easier to adopt new knowledge sources without retraining, a re-ranker or an extractive reader. 
RAG is useful to build social robots capable of generating domain-specific responses eliciting factual knowledge during HRI. 
\section{LLM Risks in HRI}
\label{sec:LLM_in_HRI_Risks}
In the previous secton, we outlined the potential of LLM in SAR design. Next, we address the risks associated with LLM.
\\ \textbf{RQ6. What risks exist in the adoption of LLM in social robot design?}
Major concerns include the perpetuation of bias and stereotypes, uncertainty over privacy, complexity, and high computation resource demands of deploying LLM.
\subsection{Bias and Stereotype Enhancement}
Bias and stereotypical thinking have been heavily studied in HRI literature \cite{10.1145/3568162.3576977, 10.1145/3171221.3171260, 10.1145/3290605.3300645, doi:10.1089/gg.2016.29002.nom, 8990011, 10.1016/j.chb.2014.05.014, wang2021useracceptancegenderstereotypes}. 
Similarly, algorithmic bias \cite{10.1145/3597307} is a significant problem with LLM. Whereas substantial efforts are being undertaken to alleviate bias related to age, sex, gender, race, religion, and the like, the training data used to train LLM contains many biases and stereotypes, posing an immense risk to the acceptability and trust of LLM.
Extra care must be taken to mitigate biases and stereotypes in behavioral training data and LLM during robot design.  
\subsection{Data Leakage}
\label{Data Leakage}
LLM users risk leaking private data to the provider of the LLM, and the LLM are susceptible to cyber-attacks \cite{Yao_2024}, which can be fatal for a social robot. 
\subsection{Long Inference Times}
Modern LLM, VLM, VLM sizes reach tens or hundreds of billions of parameters. For instance, the largest model trained in RT-2~\cite{rt22023arxiv} consists of 55B parameters. The large parameter size makes it impractical to deploy such models on standard desktop-style machines or load the LLM into the robot's computation resources for real-time robot control and dialogue. 
\section{Pathway to LLM Deployment in Social Robots}
\label{sec:Recommendations}
To harness the full potential of LLM in social robots, we recommend the following:
\textbf{Bias Mitigation:} Develop strategies to identify and mitigate bias within LLM, ensuring equitable and unbiased interactions. Moreover, principles from affirmative action can be incorporated into interactions with robots \cite{10.1145/3568162.3576977}.
\textbf{Privacy and Security:} Prioritize developing secure, privacy-preserving models that minimize data leakage and protect user information.
\textbf{Resource Efficiency:} Innovate in model compression and efficiency to enable the deployment of powerful LLM on resource-constrained platforms.
\textbf{Multimodal Learning:} Leverage the multimodal capabilities of LLM/LMM to foster richer, more natural HRI through integrating visual, auditory, and textual data.
\textbf{Ethical Guidelines:} Establish ethical guidelines for developing and deploying LLM-powered social robots, incorporating societal norms and human traits such as trust, politeness, personality, and gender considerations.

\section{Conclusion}
\label{sec:Conclusions}
Large language models (LLM) offer transformative potential for Socially Assistive Robots (SARs), enabling sophisticated and natural human-robot interactions across various applications such as education, healthcare, and entertainment.
However, the integration of LLMs in social robots presents significant technological and ethical challenges, including the risk of perpetuating biases and stereotypes, necessitating a balanced approach that addresses these issues responsibly.
This meta-study, formulated around six research questions RQ1-RQ6, lays the groundwork for responsibly integrating LLMs into SARs, contributing to the broader theme of LLMs and vision language models (VLM) in socially assistive robotics, and emphasizing the importance of training these models to understand societal norms, ethics, and human personalities before SAR deployment.
\newpage
\bibliographystyle{ACM-Reference-Format}
\bibliography{main}

\newpage
\appendix
\section{Exemplar Robots in HRI Research}
\label{Appendix:ExemplarRobotsinHRIResearch}
The Table \ref{Table:Common_Social_Robots_HRI} illustrates examples of robots covered in our meta-study.
\begin{table*}[!t]
    \footnotesize
    \centering
    \caption{A summary of the robots studied in HRI research, including the works in which the robots appeared.}
    \resizebox{\textwidth}{!}{
    \begin{tabular}{rl}
        \toprule
         \textbf{Robot} & \textbf{Exemplar Studies} \\
        \midrule
        Actroid F & \citet{10.1145/3434073.3444661} \\
        ALPHA 1P & \citet{10.5555/3523760.3523794} \\
        Alpha Mini &  \citet{10.1145/3568162.3576976} \\
        Asimo & \citet{10.1145/3434073.3444661} \\
        Astrobee & \citet{10.1145/3434073.3444644} \\
        Baxter & \citet{10.1145/3434073.3444661}, \citet{10.5555/3523760.3523792} \\
        Beam & \citet{10.1145/3568162.3576961} \\
        Bina48 & \citet{10.1145/3434073.3444661} \\
        Blossom & \citet{10.1145/3319502.3374807}, \citet{10.1145/3568162.3576987} \\
        Care-O-bot 4 & \citet{10.5555/3523760.3523777} \\
        CARMEN & \citet{10.1145/3568162.3576993} \\
        Charlie & \citet{10.1145/3568162.3576967} \\
        Create & \citet{10.1145/3434073.3444652}  \\
        Cozmo & \citet{10.1145/3319502.3374789}, \citet{10.1145/3319502.3374781}, \citet{10.1145/3319502.3374814} \\
        Double Telepresence Robot &  \citet{10.5555/3523760.3523816}, \citet{10.1145/3568162.3576961} \\
        Eva & \citet{10.1145/3319502.3374840} \\
        EMAR & \citet{10.5555/3523760.3523771} \\
        EMYS & \citet{10.5555/3523760.3523774} \\
        Face on a Globe & \citet{10.5555/3523760.3523827} \\
        Fetch & \citet{10.1145/3434073.3444658}, \citet{10.5555/3523760.3523812}, \citet{10.5555/3523760.3523815}, \citet{10.5555/3523760.3523843}, \citet{10.1145/3568162.3576998} \\
        Franka Emika Panda & \citet{10.1145/3568162.3578623} \\
        Furhat & \citet{10.1145/3319502.3374786}, \citet{10.1145/3319502.3374810}, \citet{10.1145/3434073.3444670}, \citet{10.5555/3523760.3523780}, \citet{10.5555/3523760.3523838}, \citet{10.1145/3568162.3576958}, \citet{10.1145/3568162.3578633} , \citet{10.1145/3568162.3577004} \\
        Geminoid (DK, HI, F) & \citet{10.1145/3434073.3444661} \\
        Han & \citet{10.1145/3434073.3444661} \\
        HRP-4 & \citet{10.1145/3434073.3444661} \\
        HuggieBot 2.0 & \citet{10.1145/3434073.3444656} \\
        iCub & \citet{10.1145/3319502.3374819}, \citet{10.1145/3434073.3444661}, \citet{10.1145/3434073.3444651}, \citet{10.1145/3434073.3444682}, \citet{10.1145/3434073.3444663}, \citet{10.5555/3523760.3523794} \\
        iPal & \citet{10.1145/3568162.3576960} \\
        JACO robotic arm & \citet{10.1145/3319502.3374818}, \citet{10.1145/3434073.3444667} \\
        Jibo & \citet{10.1145/3319502.3374808}, \citet{10.1145/3319502.3374794}, \citet{10.1145/3434073.3444655}, \citet{10.1145/3434073.3444671}, \citet{10.5555/3523760.3523766}, \citet{10.5555/3523760.3523769}, \citet{10.5555/3523760.3523835}, \citet{10.1145/3568162.3576968}, \citet{10.1145/3568162.3576970}, \citet{10.1145/3568162.3578625} \\
        Joy for All & \citet{10.1145/3568162.3578624} \\
        Jules & \citet{10.1145/3434073.3444661} \\
        Justin & \citet{10.1145/3434073.3444661} \\
        KIBO & \citet{10.5555/3523760.3523794} \\
        Kip1 & \citet{10.1145/2696454.2696495} \\
        Kinova Gen3 & \citet{10.1145/3568162.3576990} \\
        Kinova Jaco2 & \citet{10.1145/3568162.3578627} \\
        Kojiro & \citet{10.1145/3434073.3444661} \\
        Kuri & \citet{10.1145/3319502.3374836}, \citet{10.1145/3319502.3374821}, \citet{10.1145/3319502.3374794}, \citet{10.5555/3523760.3523771}, \citet{10.5555/3523760.3523803} \\
        LBR iiwa robot arm & \citet{10.5555/3523760.3523797} \\
        Liku & \citet{10.1145/3568162.3576970} \\
        Lovot & \citet{10.1145/3568162.3576967} \\
        Luka & \citet{10.5555/3523760.3523768} \\
        Mars Opportunity Rover & \citet{10.1145/3319502.3374794} \\
        Misty II & \citet{10.5555/3523760.3523790}, \citet{10.1145/3568162.3577003}, \citet{10.1145/3568162.3576978}  \\
        MOVO & \citet{10.1145/3568162.3576996} \\
        MyCobot & \citet{10.5555/3523760.3523811} \\
        Nadine & \citet{10.1145/3434073.3444661} \\
        NAO & \citet{10.1145/3319502.3374783}, \citet{10.1145/3319502.3374813}, \citet{10.1145/3319502.3374780},  \citet{10.1145/3319502.3374815}, \citet{10.1145/3319502.3374828}, \citet{10.1145/3319502.3374826}, \citet{10.1145/3319502.3374801}, \citet{10.1145/3319502.3374800}, \\
                                              &   \citet{10.1145/3568162.3576980}, \citet{10.1145/3319502.3374831}, \citet{10.1145/3434073.3444661}, \citet{10.1145/3319502.3374776}, \citet{10.5555/3523760.3523770}, \citet{10.5555/3523760.3523775}, \citet{10.5555/3523760.3523794}, \citet{10.5555/3523760.3523804},\\
                                              &  \citet{10.5555/3523760.3523809}, \citet{10.5555/3523760.3523821}, \citet{10.1145/3568162.3576963}, \citet{10.1145/3568162.3576957}, \citet{10.1145/3568162.3577000} \\
        Nexi MDS & \citet{10.1145/3434073.3444661} \\
        Nyokkey & \citet{10.1145/3568162.3576967} \\
        Ohmni telepresence robot & \citet{10.1145/3319502.3374785}, \citet{10.1145/3434073.3444675} \\
        OriHime, OriHimeD & \citet{10.1145/3568162.3576967}  \\
        Olly & \citet{10.1145/3568162.3576970} \\
        Ozobot Evo & \citet{10.5555/3523760.3523776} \\
        Panda & \citet{10.5555/3523760.3523799} \\
        Pepper & \citet{10.1145/3319502.3374802}, \citet{10.1145/3319502.3374797}, \citet{10.1145/3319502.3374778}, \citet{10.1145/3434073.3444673}, \citet{10.1145/3434073.3444666},  \citet{10.1145/3434073.3444661}, \citet{10.1145/3434073.3444681}, \\
                                                & \citet{10.5555/3523760.3523822}, \citet{10.5555/3523760.3523830}, \citet{10.5555/3523760.3523831}, \citet{9889624}, \citet{10.1145/3568162.3576959}, \citet{10.1145/3568162.3576999}, \citet{10.1145/3568162.3576967}, \citet{10.1145/3568162.3576971} \\
        Philip K. Dick & \citet{10.1145/3434073.3444661} \\
        PR2 & \citet{10.1145/3434073.3444687}, \citet{10.5555/3523760.3523812} \\
        QTRobot & \citet{10.5555/3523760.3523805}, \citet{10.1145/3568162.3577003} \\
        Zeno R25 & \citet{10.1145/3319502.3374803} \\
        REEM & \citet{10.5555/3523760.3523793} \\
        RoboHon & \citet{10.5555/3523760.3523794}, \citet{10.1145/3568162.3576967} \\
        Robovie-R3/R2 & \citet{10.1145/3319502.3374812}, \citet{10.1145/3319502.3374830}, \citet{10.1145/3319502.3374833}, \citet{10.1145/3434073.3444679}, \citet{10.1145/3434073.3444674} \\
        Roboy & \citet{10.5555/3523760.3523794}  \\
        Romi & \citet{10.1145/3568162.3576967} \\
        Sawyer & \citet{10.1145/3319502.3374838}, \citet{10.1145/3319502.3374791}, \citet{10.1145/3319502.3374839}, \citet{10.1145/3434073.3444649}, \citet{10.1145/3434073.3444664}, \citet{10.1145/3568162.3576967} \\
        Scitos G5 & \citet{10.5555/3523760.3523800}, \citet{10.1145/3568162.3578627} \\
        RP-7 & \citet{10.1145/3319502.3374775} \\
        Stevie (version II) & \citet{10.1145/3319502.3374834} \\
        Stretch & \citet{10.1145/3568162.3576994} \\
        Showa Hanako & \citet{10.1145/3434073.3444661} \\
        Shutter & \citet{10.1145/3568162.3576986} \\
        Shybo & \citet{10.5555/3523760.3523837} \\
        SocibotMini & \citet{10.1145/3319502.3374787} \\
        Sophia & \citet{10.1145/3434073.3444687} \\
        Sota & \citet{10.5555/3523760.3523829}, \citet{10.1145/3568162.3577005}   \\
        TIAGo & \citet{10.1145/3434073.3444680} \\
        The Greeting Machine & \citet{8525516} \\
        Thymio & \citet{10.1145/3434073.3444652} \\
        TurtleBot 2 & \citet{10.1145/3434073.3444676}, \citet{10.5555/3523760.3523789}, \citet{10.5555/3523760.3523823} \\
        Twendy One & \citet{10.1145/3434073.3444661} \\
        UR (5,10) robot arm & \citet{10.1145/3319502.3374841}, \citet{10.1145/3319502.3374820}, \citet{10.1145/3319502.3374829}, \citet{10.1145/3434073.3444647}, \citet{10.1145/3319502.3374779}, \citet{10.5555/3523760.3523787}, \citet{10.1145/3568162.3576990}, \citet{10.1145/3568162.3576969}, \citet{10.1145/3568162.3576964} \\
        Vector & \citet{10.1145/3434073.3444659}, \citet{10.1145/3434073.3444665} \\
        WAM & \citet{10.1145/3568162.3577002} \\
        YOLO & \citet{10.1145/3319502.3374817}, \citet{10.1145/3434073.3444650} \\
    \bottomrule
    \end{tabular}
    }
    \label{Table:Common_Social_Robots_HRI}
\end{table*}

\section{Applications of Robots}
\label{Appendix:ApplicationsOfRobots}
Examples of major robot applications covered in our meta-study are shown in Table 
\begin{table*}[!t]
\footnotesize
\centering
\caption{Exemplar applications of socially assistive robots by sector.}
\resizebox{\textwidth}{!}{
\begin{tabular}{rl}
\toprule
\textbf{Research Work} & \textbf{Details of Applications \& Robots used }  \\
\midrule
                & \textbf{Healthcare:}     \\
\citet{10.1145/3319502.3374836} & deployed \texttt{Kuri} to define individual-specific and interactive therapies for mild cognitive impairment (MCI) patients. \\
\citet{10.1145/3568162.3576993} & introduced \texttt{CARMEN}, Cognitively Assistive Robot for Motivation and Neurorehabilitation, to deliver neuro-rehabilitation to people with mild cognitive impairment (PwMCI). \\
\citet{10.5555/3523760.3523771} & introduced robot prototypes based on \texttt{Kuri, EMAR} to deliver cognitive training for people with mild cognitive impairment (PwMCI). \\
\citet{10.1145/3319502.3374797} & included \texttt{Pepper} in an interactive game to rehabilitate stroke patients. \\
\citet{10.1145/3319502.3374775} & experimented with remote diagnosis of stroke patients. \\
\citet{10.1145/3319502.3374840} & studied the impact of \texttt{Eva} on reducing behavioral and psychological symptoms of dementia (BPSD) for people with dementia. \\
\citet{10.1145/3319502.3374826} & developed design patterns with \texttt{Nao} for better child engagement to foster rehabilitation during pediatric care. \\
\citet{10.1145/3319502.3374818} & investigated the possibility of robot-assisted feeding for people with mobility impairments. \\
\citet{10.1145/3434073.3444671} & used \texttt{Jibo} to deliver daily positive psychology sessions to undergraduate students. \\
\citet{10.1145/3568162.3577003} & deployed  \texttt{QT robot} and \texttt{Misty II} as robotic coaches to deliver four positive psychology exercises aimed at preserving the mental well-being of employees at an organization. \\
\citet{10.1145/3568162.3578625} & deployed \texttt{Jibo} to study the mental and psychological well-being of people aged 18 to 83. \\
\citet{10.1145/3568162.3576987} & deployed \texttt{Blossom} to investigate the impact of text-to-speech (TTS) and human voices in mindfulness meditation. \\
\citet{10.5555/3523760.3523766} & deployed \texttt{Jibo} to study how people with Autism Spectrum Disorders (ASD) can cope with interactions. \\
\citet{10.5555/3523760.3523770} & used \texttt{Nao} to study autism amongst children. \\
\citet{10.1145/3568162.3576963}  & used \texttt{Nao} to study Autism Spectrum Disorder (ASD) amongst kids. \\
\citet{10.5555/3523760.3523838} & employed \texttt{Furhat} for peripartum depression (PPD) screening. \\
\citet{10.1145/3568162.3576960} & introduced emotion gestures into \texttt{iPal} drawing increased engagement and lower anxiety for children during vaccination. \\
\citet{10.1145/3568162.3576971} & developed a contact-free solution based on \texttt{Pepper} and the \textit{Transformer}\citet{vaswani2023attention} for automated delirium detection in hospital wards using a \\
                                & robotic implementation of the Confusion Assessment Method for the Intensive Care Unit (CAM-ICU). \\
\citet{10.1145/3568162.3576994} & deployed \texttt{Stretch} and explored ways for interruption-mitigation and reorientation methods for mobile telemanipulator robots (MTRs) in emergency departments (ED). \\
\midrule
                & \textbf{Education:}  \\
\citet{10.1145/3319502.3374803} & deployed \texttt{Zeno R25} to study the interactive behaviour between students and robots. \\
\citet{10.1145/3319502.3374813} & investigate the potential of school children to learn to hand-write the Kazakh alphabet by teaching the \texttt{Nao} robot how to write Kazakh. \\
\citet{10.1145/3319502.3374787} & used \texttt{SocibotMini} to study how social robots can influence learning how to play musical instruments. \\
\citet{10.1145/3319502.3374815} & deployed \texttt{Nao} to study the impact of the robot gestures on school children's learning. \\
\citet{10.1145/3319502.3374817} & leveraged \texttt{YOLO} to understand the impact of a robot on children's creativity. \\
\citet{10.1145/3319502.3374822} & found out that the presence of a tutee robot improves the predictability of children's affective displays during vocabulary learning.  \\
\citet{10.1145/3319502.3374792} & proposed a new platform to facilitate interaction between child, parent, and robot, and such interactions are beneficial to the toddler's early development and learning. \\
\citet{10.1145/3434073.3444670} & investigated the ability of a native and a second-language learner to play a language skill-dependent game. \\
\citet{10.5555/3523760.3523768} & leveraged \texttt{Luka} to helps toddlers to read. \\
\citet{10.5555/3523760.3523774} & used \texttt{EMYS} to stimulate creativity among school children. \\
\citet{10.5555/3523760.3523787} & used \texttt{UR5e} to teach adults how to build an electric circuit under \textit{peer} and \textit{tutor} roles. \\
\citet{10.5555/3523760.3523837} & used \texttt{Shybo} to investigate the possibility of promoting critical thinking in school children. \\
\citet{10.1145/3568162.3576957} & used \texttt{Nao} to study concrete design specifications for creating a more engaging and effective learning experience by intertwining the social behaviors of robot math tutor with the math task. \\
\citet{10.1145/3568162.3576968} & deployed a \texttt{Jibo} robot companion to encourage and motivate children to explore during storybook reading, enabling children's literacy learning. \\
\citet{10.1145/3568162.3578633} & deployed a \texttt{Furhat} robot tutor to teach native-Dutch-speaking participants to pronounce Japanese words.  \\
\midrule
                & \textbf{Entertainment:} \\
\citet{10.1145/3319502.3374780} & studied the humorous part of robots by enabling \texttt{NAO} to tell jokes to an audience. \\
\citet{10.1145/3319502.3374809} & the robot is programmed to take photos in portrait mode. \\
\citet{10.1145/3319502.3374837} & a wearable robotic device that creates a cyborg character is involved in a dance performance. \\
\citet{10.1145/3434073.3444682} & developed a framework that enabled \texttt{iCub} to autonomously lead an entertaining and effective human-robot interaction based on the real-time reading of a biometric feature from the players. \\
\citet{10.5555/3523760.3523834} & proposed a new understanding of improvisation based on rules that shape robot movement and behavior, leading to increased engagement and responsiveness in a dance performance. \\
\midrule
                & \textbf{Hospitality and Services:} \\
\citet{10.1145/3319502.3374795} & developed a robot to study the impact of \textit{co-embodiment} and \textit{re-embodiment} in the services domain such as Quick Care Clinic, Canton Department Store, Homestead Inn. \\
\citet{10.5555/3523760.3523831} & trained \texttt{Pepper} to behave appropriately as a waiter in a restaurant and to respond to customer requests. \\
\citet{10.5555/3523760.3523829} & investigated the influence of several forms of social presence of teleoperated robots on customer behavior. \\
\citet{10.1145/3568162.3576967} & studied environments where several robots were deployed as service robots in \textit{robot cafes}. \\
                               & The robots included \texttt{Lovot, Pepper, OriHime, OriHimeD, Sawyer, Romi, Charlie, RoBoHoN, and Nyokkey}. \\
\citet{10.1145/3568162.3577005} & found that deployment of service robots (named \texttt{Sota}) in pairs resulted in increased sales at a bakery store.  \\
\citet{10.1145/3568162.3576984} & deployed a delivery robot to dispatch food and "convenience store" products to customers at various locations and investigated the impact of groups towards acceptance and trust of robots.\\
\midrule
                & \textbf{Tele-operations and Telepresence:} \\
\citet{10.1145/3568162.3576961} & studied the lived experience of participating in hybrid spaces through a telepresence robot. The robots used in this study were the \texttt{Double} by Double Robotics and the \texttt{Beam} by Suitable Technologies \\
\bottomrule
\end{tabular} 
}
\label{Table:ApplicationsofAssistiveRobots} 
\end{table*}

\end{document}